\documentclass{IEEEtran}
\usepackage{cite}
\usepackage{amsmath,amssymb,amsfonts}
\usepackage{graphicx}
\usepackage{textcomp}
\usepackage{algorithm}
\usepackage{algorithmic}
\usepackage{multirow}
\renewcommand{\algorithmicrequire}{ \textbf{Input:}}     
\renewcommand{\algorithmicensure}{ \textbf{Initialize:}} 
\def\BibTeX{{\rm B\kern-.05em{\sc i\kern-.025em b}\kern-.08em
    T\kern-.1667em\lower.7ex\hbox{E}\kern-.125emX}}
\begin{document}
\title{SiftingGAN: Generating and Sifting Labeled Samples to Improve the Remote Sensing Image Scene Classification Baseline \emph{in vitro}}
\author{Dongao Ma, Ping Tang, and Lijun Zhao
\thanks{This work was supported by the Strategic Priority Research Program of the Chinese Academy of Sciences under Grant XDA19080301, the National Natural Science Foundation of China under Grant 41701397 \& 41701399, and the Major Project of High Resolution Earth Observation System of China under Grant 03-Y20A04-9001-17/18.  (\emph{Corresponding author: Lijun Zhao}.)}
\thanks{D. A. Ma is with the Institute of Remote Sensing and Digital Earth, Chinese Academy of Sciences, Beijing 100101, China, and also with the University of Chinese Academy of Sciences, Beijing 100049, China (e-mail: mada@radi.ac.cn).}
\thanks{P. Tang and L. J. Zhao are with the Institute of Remote Sensing and Digital Earth, Chinese Academy of Sciences, Beijing 100101, China (e-mail: tangping@radi.ac.cn; zhaolj01@radi.ac.cn).}
}

\maketitle

\begin{abstract}

Lack of annotated samples greatly restrains the direct application of deep learning in remote sensing image scene classification. Although researches have been done to tackle this issue by data augmentation with various image transformation operations, they are still limited in quantity and diversity. Recently, the advent of the unsupervised learning based generative adversarial networks (GANs) bring us a new way to generate augmented samples. However, such GAN-generated samples are currently only served for training GANs model itself and for improving the performance of the discriminator in GANs internally (\emph{in vivo}). It becomes a question of serious doubt whether the GAN-generated samples can help better improve the scene classification performance of other deep learning networks (\emph{in vitro}), compared with the widely used transformed samples. To answer this question, this paper proposes a SiftingGAN approach to generate more numerous, more diverse and more authentic labeled samples for data augmentation. SiftingGAN extends traditional GAN framework with an Online-Output method for sample generation, a Generative-Model-Sifting method for model sifting, and a Labeled-Sample-Discriminating method for sample sifting. Experiments on the well-known AID dataset demonstrate that the proposed SiftingGAN method can not only effectively improve the performance of the scene classification baseline that is achieved without data augmentation,  but also significantly excels the comparison methods based on traditional geometric/radiometric transformation operations.
\end{abstract}

\begin{IEEEkeywords}
Deep learning, generative adversarial networks, scene classification, data augmentation.
\end{IEEEkeywords}

\section{Introduction}
\label{sec:Introduction}
Remote sensing image scene classification\cite{Zhao2017Land} plays a significant role in a broad range of application, such as urban planning, environment monitoring, and land-resource management. The main problem of scene classification is constructing robust and discriminative feature descriptors. In recent years, deep learning based methods, such as deep convolutional neural networks (DCNN)\cite{Zhao2018Analysis} and recurrent neural networks (RNN)\cite{huang2018long,Wang2018Scene}, have become the state-of-the-art solutions for scene classification. The deep learning algorithms can automatically learn the high-level semantic features and have shown great strength in remote sensing field. However, this kind of methods require extremely large datasets with rich contents to train a model for capturing the essential features of scenes, otherwise the capability of deep networks will be restricted and the ideal accuracy result will be hard to achieve. However, it is difficult to obtain such enormous amount of labeled remote sensing images since the data annotation is time-consuming and expensive. To tackle this issue, a lot of efforts have been made, where data augmentation is the easiest way and the most common used method. It applies image transformation operations to augment the dataset\cite{Yu2017Deep}. But there is limitation in the enhancement of quantity and diversity.

Generative adversarial networks (GANs) \cite{Goodfellow2014Generative}, first proposed by Goodfellow \emph{et al.} in 2014, is a generative model which consists of a generator and a discriminator. The generator captures the data distribution and generates simulated samples. The discriminator estimates the probability that a sample came from the training data rather than the generator. The purpose of GANs is to estimate the potential distribution of data through a two-player min-max game. The discriminator’s capability of data feature extraction is upgraded via the adversarial training process, thus GANs can be used to unsupervised representation learning with unlabeled data. Based on this, many researchers have noticed that GANs could be an excellent solution for the lack of labeled remote sensing data. Lin \emph{et al.} \cite{Lin2017MARTA} proposed a multiple-layer feature-matching GANs model based on deep convolutional GANs (DCGAN) to learn the unsupervised representation of remote sensing images. Xu \emph{et al.} \cite{Xu2018Remote} replaced the Rectified Linear Unit (ReLU) and batch normalization of DCGAN with the Scaled Exponential Linear Units (SELU) for a supervised scene classification. Not only in scene image classification, GANs has also been applied in hyperspectral image classification \cite{Zhu2018Generative,He2017Generative}. However, although these works mentioned that they utilized the newly generated samples for training GANs, such usage of samples only performs the improvement in the discriminator of GANs internally (\emph{in vivo}). 

Thus there exist two questions, one is whether the GAN-generated samples can be used in other deep learning frameworks out of GANs (\emph{in vitro}) to attain a higher classification accuracy, and the other one is that whether the simulated samples generated by GANs perform better than the transformed samples produced by traditional operations. These questions still remain unknown for remote sensing scene classification. As opposed to ``\emph{in vivo}", the term ``\emph{in vitro}" \cite{Zheng2017Unlabeled} in this paper refers to \emph{generating samples with a GAN and feeding them into a separate DCNN along with “regular” training data}.

In this work, we propose a GAN based approach namely \emph{SiftingGAN}, to generate and sift labeled samples to augment sample dataset. Then we conduct \emph{in vitro} experiments to prove that the GAN-generated samples perform better than the traditional ways as a data augmentation method. 

Compared with traditional GAN framework, in SiftingGAN, firstly, we change the samples output mode to produce more numerous, diverse samples. We propose an \emph{Online-Output} mode to update the networks continuously and generate samples synchronously rather than terminating the training process when the model is well trained. This is equivalent to utilizing plenty of models to produce samples and can guarantee the diversity of generated samples.

Secondly, we add two sifting processes in generator and discriminator respectively to output the best part of samples. One sifting process is \emph{Generative-Model-Sifting}, it uses the generator’s loss value to pick out the better models with low losses. These models are able to generate better batches of images containing every class, because the lower generator loss reflects the smaller difference between the overall generative distribution and the training-set distribution. The other sifting process is \emph{Labeled-Sample-Discriminating}, it uses discriminator to estimate the probability that a labeled generated sample is discriminated as a real one. This process can sift out the samples that own high probability value.

The paper is organized as follows. Section \ref{sec:Methodology} describes the proposed SiftingGAN in detail. Section \ref{sec:Experiments} introduces the experiments in vitro and presents experimental results. Conclusions are summarized in Section \ref{sec:Conclusion}.

\begin{figure}[!t]
\centerline{\includegraphics[width=\columnwidth]{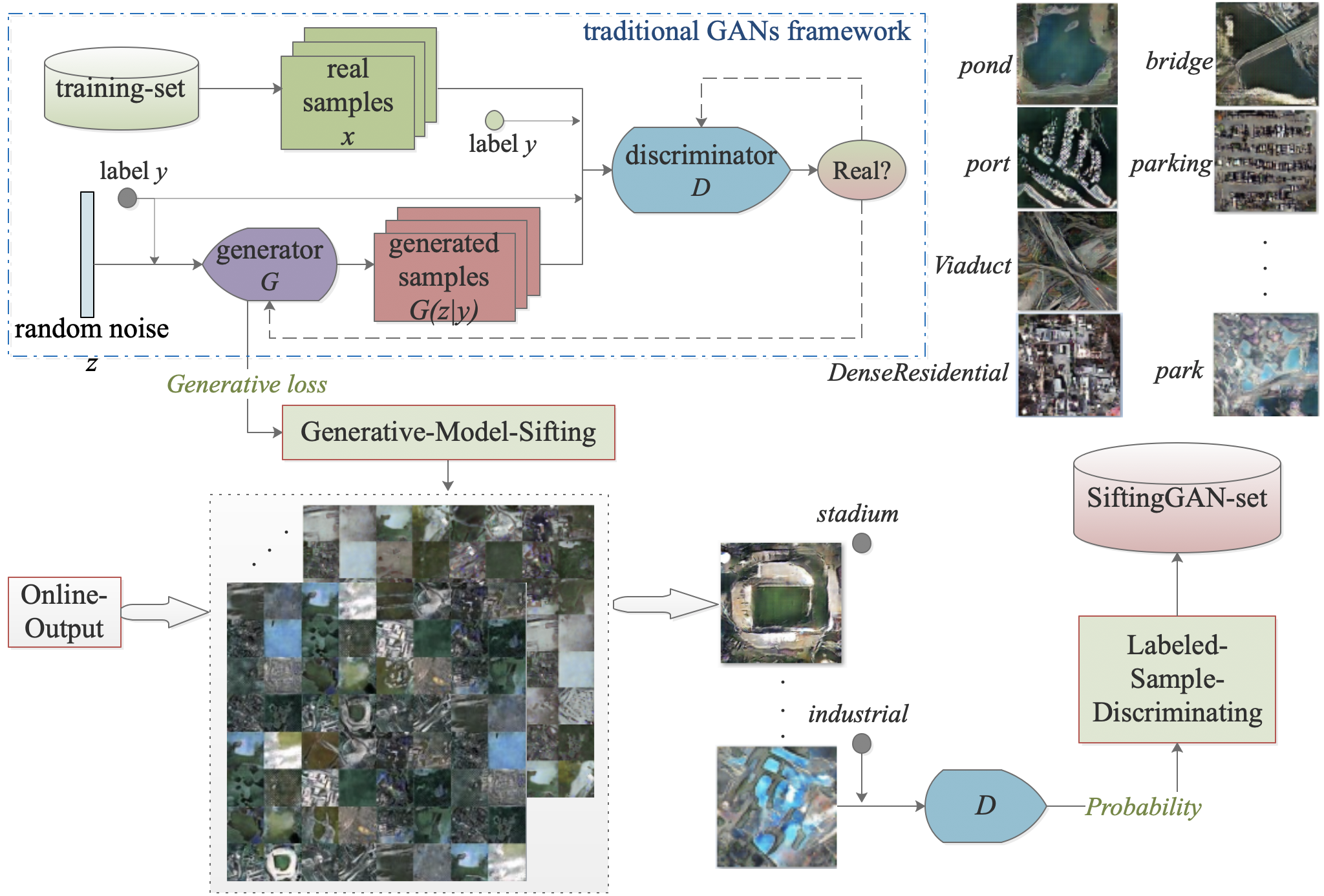}}
\caption{The architecture of the proposed SiftingGAN. The traditional GANs framework is extended with a different output mode and two sifting processes.}
\label{fig1}
\end{figure}

\section{Methodo{}logy}
\label{sec:Methodology}
\subsection{Generative Adversarial Networks}
GANs proposed by Goodfellow \emph{et al.} consist of two networks: a generator $G$ and a discriminator $D$. The generator aims to learn a generative distribution $p_g$ as close as possible to the data distribution $p_{data}$, and then generates new data samples from $p_g$ by mapping from a prior noise distribution $p_z$(z) to data space as $G(z;\theta_g)$. The discriminator $D(x;\theta_d)$, is to estimate the distance between $p_g$ and $p_{data}$ by computing the probability that a sample $x$ came from the training data rather than $p_g$. $G$ and $D$ are both trained simultaneously, and their parameters $\theta_g$ and $\theta_d$ are updated alternately. The objective function $V(D,G)$ of GANs is represented as:
\begin{multline}
\min_{G}\max_{D}V(D,G)=\mathbb{E}_{x \sim p_{data}(x)}[logD(x)]+\\
\mathbb{E}_{z \sim p_z(z)}[log(1-D(G(z)))]
\end{multline}

Previous researches\cite{Lin2017MARTA,Xu2018Remote,Zhu2018Generative,He2017Generative} followed the classic usage of GANs and performed the representation learning \emph{in vivo}. Their main purpose was to train a GANs model for classification task. They focused on the discriminative network which can learn a good data representation during the adversarial process. In the mean time, the generative network generated simulated samples along with real samples to train the discriminative network. The generated samples were used inside GANs.

\subsection{The Proposed SiftingGAN}
Fig. 1 shows the architecture of the proposed approach. In this architecture, the traditional GANs framework is extended with an \emph{Online-Output} mode and two sifting processes, \emph{Generative-Model-Sifting} and \emph{Labeled-Sample-Discriminating}, aiming to generate more numerous, more diverse, and more authentic annotated samples as a data augmentation method used for other models \emph{in vitro}. We call this architecture ``\emph{SiftingGAN}''. In this work, we adopt conditional GANs (cGANs)\cite{Mirza2014Conditional} to integrate label information into both generator and discriminator, which made the data generation process controllable with class label $y$. Moreover, we combine cGANs with DCGAN \cite{Radford2015Unsupervised} for better extraction of spatial information in remote sensing scene images. The objective function in this work is:
\begin{multline}
\min_{G}\max_{D}V(D,G)=\mathbb{E}_{x \sim p_{data}(x)}[logD(x \mid y)]+\\
\mathbb{E}_{z \sim p_z(z)}[log(1-D(G(z \mid y)))]
\end{multline}

\subsubsection{Online-Output}
\begin{figure}[!t]
\centerline{\includegraphics[width=\columnwidth]{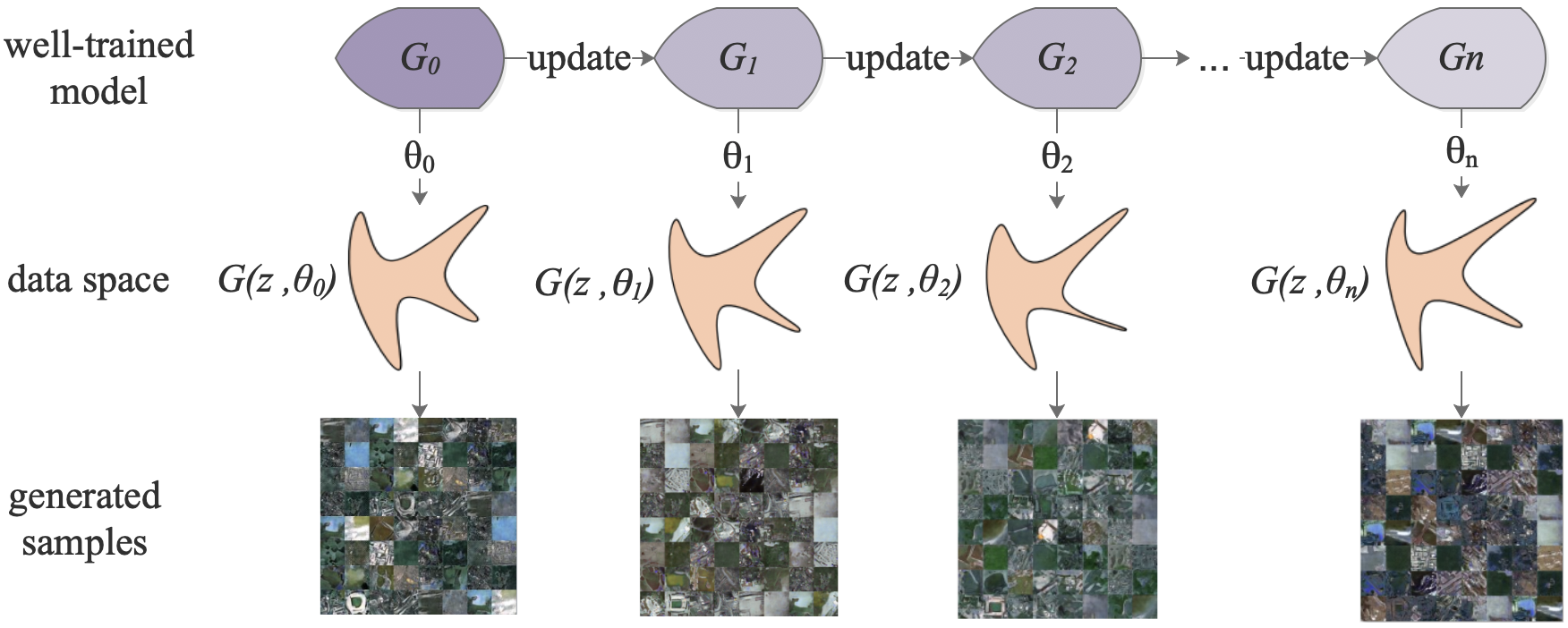}}
\caption{The proposed \emph{Online-Output} is equivalent to utilizing plenty of models to generate samples.}
\label{fig2}
\end{figure}

Normally, the training process is terminated when the model is well trained. The generation of random noise vector $z$ is unlimited, so that the well-trained $G$ can produce countless samples. However, since the well-trained $G$ with fixed network parameters $\theta_g$ will always output samples in a data space $G(z;\theta_g)$ which is decided by $\theta_g$, the diversity of generated dataset as well as the increment of novel information content will be limited in $G(z;\theta_g)$. Thus we propose an \emph{Online-Output} mode to cover such shortage. At first, we train the GANs model in the traditional way from the beginning. Then the \emph{Online-Output} begins to work when the training of both $G$ and $D$ converges. At the same time, we continue to feed samples to GANs and update the networks continuously rather than terminating the training. Every time the model parameters are updated, the generator $G$ with new parameters $\theta_g'$ will output fresh samples in a novel data space $G(z;\theta_g')$. As shown in Fig. \ref{fig2}, \emph{Online-Output} is equivalent to utilizing plenty of models to produce samples, which guarantees the diversity of generated samples.

\subsubsection{Generative-Model-Sifting}
\emph{Online-Output} can produce more numerous and more diverse samples by using lots of different models. But not every model can output fine samples. To tackle this issue, we propose a \emph{Generative-Model-Sifting} method to pick the better models. The purpose of $G$ is to learn the data distribution and to produce samples that fool $D$. Therefore, $G$ expects to maximize $D(G(z \mid y))$ or equivalently minimize $1-D(G(z \mid y))$. And the loss function of $G$ is:
\begin{equation}
\mathcal{L}(G)=\mathbb{E}_{z \sim p_z(z)}[log(1-D(G(z \mid y)))]
\end{equation}
The more fake samples that $D$ discriminates to be real, the less loss value that $G$ can get. This reflects that the model can generate better samples. Thus we can use the loss function of $G$ as the criterion to estimate the discrepancy between $p_g$ and $p_{data}$. In Algorithm \ref{algo1}, the general steps of \emph{Generative-Model-Sifting} process is given.
\begin{algorithm}
	\caption{Generative-Model-Sifting method}
	\label{algo1}
	\begin{algorithmic}[1]
	\renewcommand{\algorithmicrequire}{\textbf{Input:}} 
	\renewcommand{\algorithmicensure}{\textbf{Output:}} 
	\REQUIRE A batch of random noise vectors $\{Z_i\}_{i=1}^n=(Z_1,Z_2,...,Z_n)$ with class labels $\{y_i\}_{i=1}^n \in \{Y^1,Y^2,...,Y^k\}$, $n$ denotes the batch size and $k$ denotes the number of dataset classes.
	\ENSURE A batch of GAN-generated samples or $null$.
		\STATE Given a well-trained generative model $G_j$ from the \emph{Online-Output} method, generate a batch of samples: \\$\{I_i,y_i\}_{i=1}^n=G_j(\{Z_i,y_i\}_{i=1}^n)$;
		\STATE Feed the batch of samples $\{I_i,y_i\}_{i=1}^n$ to $D_j$ and estimate the probability values that the samples are real: $D_j(\{I_i,y_i\}_{i=1}^n)$;
		\STATE Calculate the loss value: \\$\mathcal{L}(G_j)=\frac{1}{k}\sum_{i=1}^n log[1-D(\{I_i,y_i\}_{i=1}^n)]$;
		\STATE Set a loss threshold value ${\tau}$, if $\mathcal{L}(G_j) < {\tau}$ , then output the batch of samples $\{I_i,y_i\}_{i=1}^n$ generated by $G_j$.
		
	\end{algorithmic}
\end{algorithm}

\subsubsection{Labeled-Sample-Discriminating}
After the \emph{Generative-Model-Sifting} process, the better part of the models produced by \emph{Online-Output} is picked out for production. These models not only perform better on diversity, but also have stronger ability to generate samples that with higher probability to be authenticated, nevertheless, they may not be able to ensure quality of every generated sample in a batch. Thus we propose a \emph{Labeled-Sample-Discriminating} method using the well-trained $D$ to discriminate every single labeled generated sample. The duty of $D$ is to estimate the probability that a sample came from the training data rather than $G$. When we integrate the class information into GANs, $D$ estimates the conditional probability that one labeled sample came from a certain class of the training data. The general steps of \emph{Labeled-Sample-Discriminating} method is given in Algorithm \ref{algo2}.
\begin{algorithm}
	\caption{Labeled-Sample-Discriminating method}
	\label{algo2}
	\begin{algorithmic}[1]
	\renewcommand{\algorithmicrequire}{\textbf{Input:}} 
	\renewcommand{\algorithmicensure}{\textbf{Output:}} 
	\REQUIRE A labeled sample $(I_s,y_s)\in\{I_i,y_i\}_{i=1}^n$ from the batch of samples generated by Algorithm \ref{algo1}.
	\ENSURE A labeled sample or $null$. 
		\STATE Feed the input sample $s=(I_s,y_s)$ to a well-trained disciminative model $D_j$ ;
		\STATE Get a probability value $D_j(s)$ that the sample $s$ is discriminated as a real one;
		\STATE Set a probability threshold value ${\rho}$, if $D_j(s) > {\rho}$, then output the sample $s$ and put it into the SiftingGAN-set.
	\end{algorithmic}
\end{algorithm}

\section{Experiments}
\label{sec:Experiments}

\begin{figure}[!t]
\centerline{\includegraphics[width=\columnwidth]{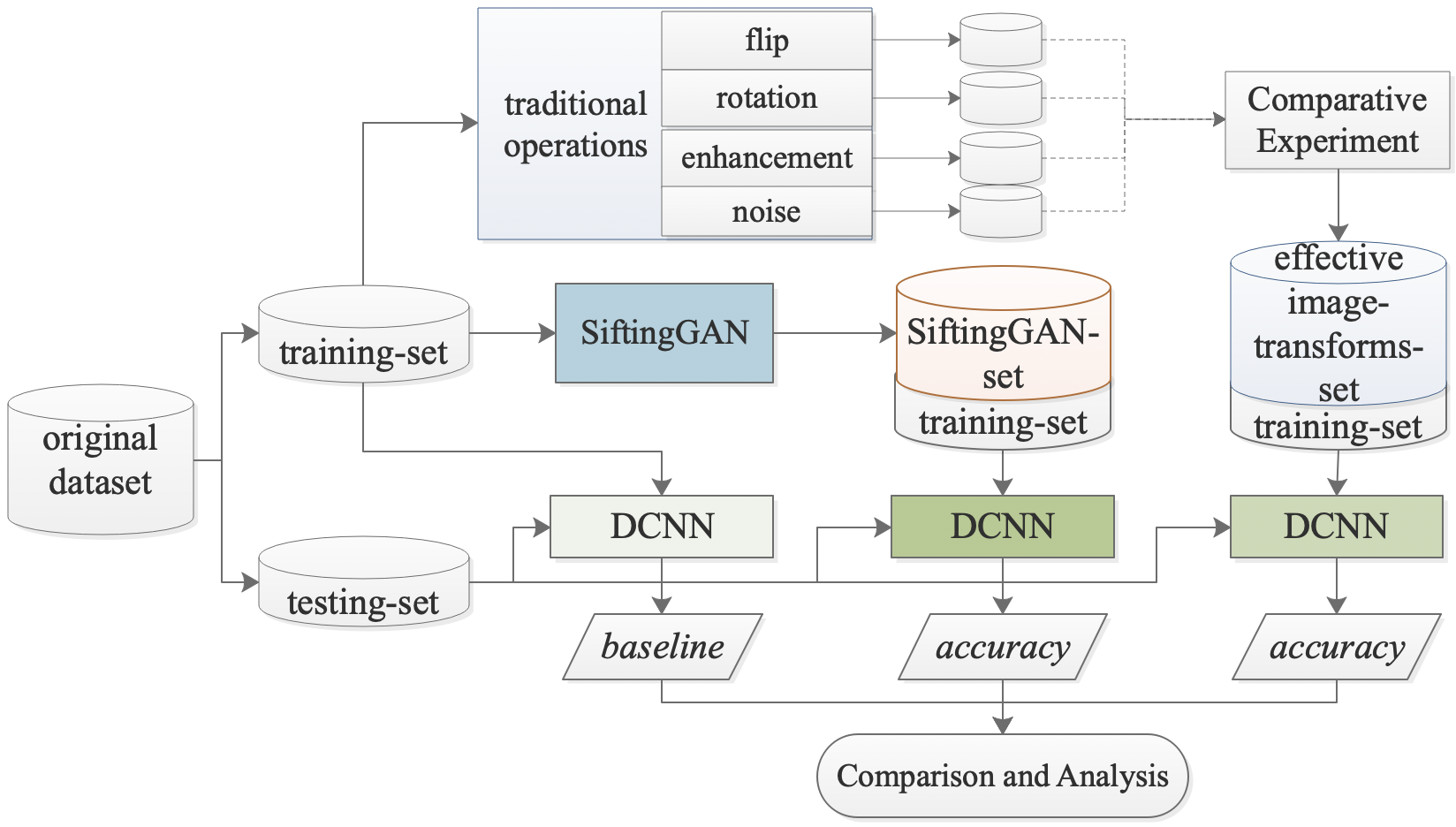}}
\caption{Overview of the \emph{in vitro} experiments in this work.}
\label{fig3}
\end{figure}

As shown in Fig. \ref{fig3}, an original dataset is separated into training-set and testing-set. Based on the training-set, three kinds of datasets can be obtained: the training-set itself ($T_{origin}$), a training set augmented by the SiftingGAN-set ($T_{SifGAN}$), and a training set augmented by the image transforms based datasets ($T_{transf}$). To verify the effectiveness of the proposed SiftingGAN, a baseline experiment is conducted without data augmentation, and two groups of ``\emph{in vitro}'' experiments are performed with data augmentation. In baseline experiment, $T_{origin}$ is used independently to train a DCNN model as the classification baseline model. In the ``\emph{in vitro}'' experiments, the DCNN models  are trained by $T_{SifGAN}$ and $T_{transf}$ separately, whose classification accuracies on testing-set are compared. In addition, to generate $T_{transf}$, different transformation operations including geometric and radiometric based methods are compared and the most effective data augmentation methods are selected for comparison with the proposed SiftingGAN method.
In this work, the well-known AID dataset is used as the original dataset and the VGGNet16 is chosen as an ``\emph{in vitro}'' DCNN framework.

\subsection{AID dataset and VGGNet16}
The Aerial Image Dataset (AID)\cite{Xia2017AID} is a large-scale dataset for aerial scene classification. It has a number of 10000 images within 30 classes. The size of each aerial image is fixed to be $600 \times 600$ pixels, and the pixel-resolution changes from about 8 meters to about half a meter.
In \cite{Xia2017AID}, different deep learning approaches were compared on several datasets. Overall, the fine-tuned VGGNet16\cite{Simonyan2014Very} method has simpler architecture and better performance on scene classification task. Thus the VGGNet16 is chosen as an ``\emph{in vitro}'' DCNN framework and a pre-trained model is utilized for fine-tuning on $T_{origin}$, $T_{SifGAN}$ and $T_{transf}$ respectively to obtain different models for comparison.

\subsection{Implementation Details}
The AID dataset is randomly split into 50\% as the training-set and 50\% as the testing-set. Due to the limitation of our computing capability, the samples in the dataset are down-sampled to a size of $256 \times 256$ pixels. To quantitatively evaluate the performances of proposed SiftingGAN and traditional image-transforms based methods, different VGGNet16 models trained with $T_{SifGAN}$ and $T_{transf}$ are tested on the testing-set, and the classification result was evaluated by average overall classification accuracy from five runs.

\subsubsection{The fine-tuned VGGNet16 baseline}
VGGNet16 has 13 convolutional layers, 5 pooling layers, and 3 fully-connected layers. In our expriments, the output layer is initialized anew while the other layers of the pre-trained VGGNet16 model are retained. To avoid that the inadequate-trained output layer discourages the front layers, we only train the output layer at first. Then we fine-tune the convolutional layers and the fully-connected layers with different learning rate of 0.0001 and 0.001 respectively. At last the fine-tuned model is used for scene classification on the testing-set. The overall accuracy without data augmentation is the evaluating baseline of our experiments.

\subsubsection{Image transformation based data augmentation}
We apply traditional image-transforms based data augmentation methods using geometric and radiometric transformation operations, including flip (horizontally, vertically, $45^{\circ}$ diagonally, $135^{\circ}$ diagonally), rotation ($90^{\circ}$, $180^{\circ}$, $270^{\circ}$), image enhancement (Laplacian filter, Gamma transform, histogram equalization), noise (Gaussian-distributed, Poisson-distributed, salt and pepper additive noise), to augment the training-set.

\subsubsection{SiftingGAN training and samples generating}
\begin{figure}[!t]
\centerline{\includegraphics[width=\columnwidth]{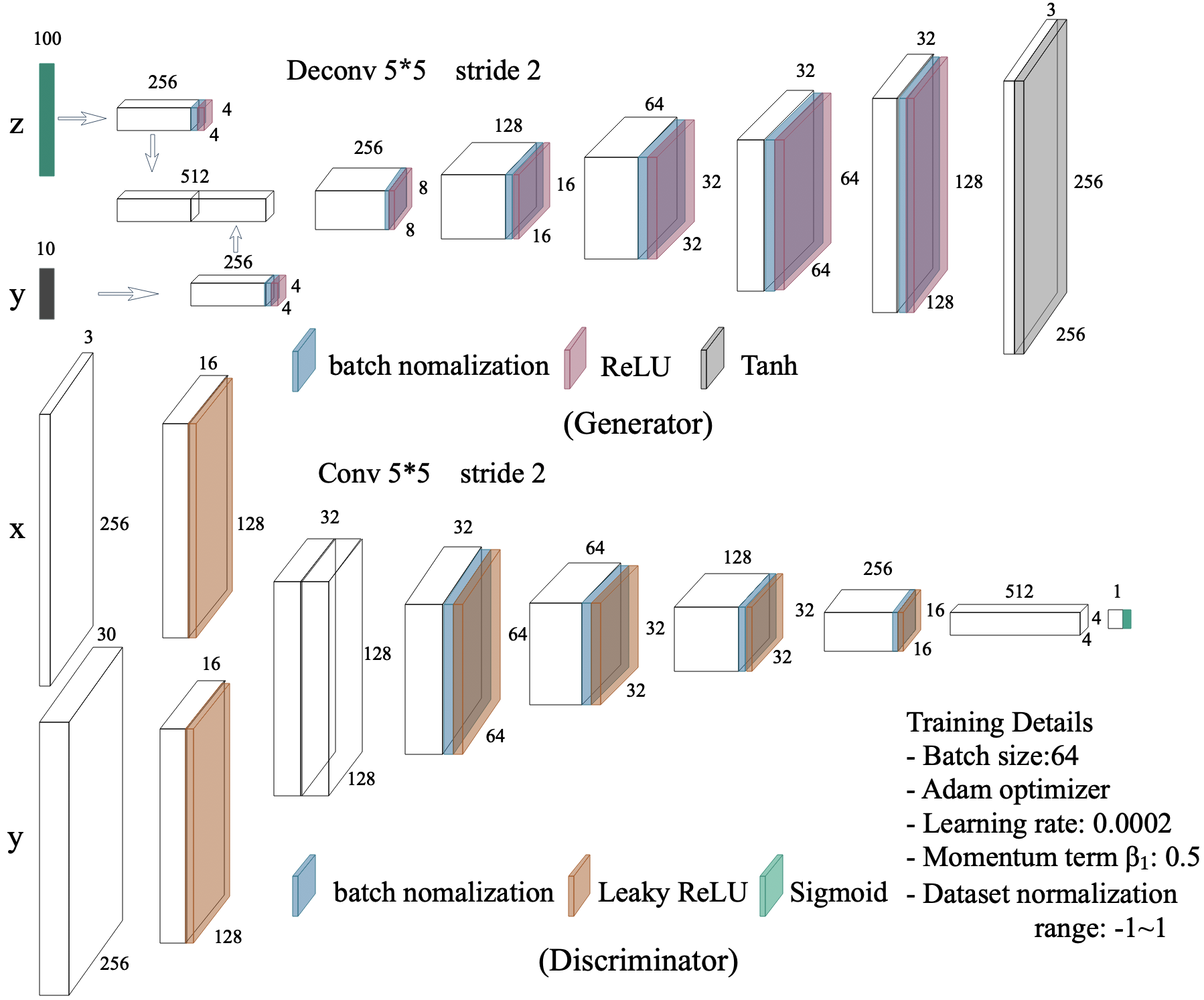}}
\caption{Network architectures of generator and discriminator in this work.}
\label{fig4}
\end{figure}

The proposed SiftingGAN is implemented with TensorFlow. It supports inputing extra label information and outputing $256 \times 256$ annotated images. 
More details of the network architectures are shown in Fig.\ref{fig4}.
To maintain learning balance between $G$ and $D$, we double the learning rate of $G$ after 25k iterations and double again after 50k iterations. Then we trigger the Online-Output and the two sifting processes to generate labeled samples. We set the loss threshold $\tau$ to 1.0 and the probability threshold $\rho$ to 0.9.

\subsubsection{Model training with data augmentation}
The samples in SiftingGAN-set come from the overall generative distribution. Although each one has its own label, it may contain the information from other classes. One class label cannot be regarded as ground truth. Thus we adopt Label Smoothing Regularization (LSR)\cite{Zheng2017Unlabeled} method to take the none-ground truth classes into account. When we use SiftingGAN-set or image-trainsforms-sets for data augmentation, we randomly take samples from it in multiples (5k, 10k, ..., 40k) of the training-set number (5k). The real samples in training-set and image-trainsforms-sets are given one-hot ground truth labels, and the simulated samples in SiftingGAN-set are assigned LSR regularized labels. Every batch of input consists of such two part (original/augmented) samples in a certain ratio (1:1, 1:2, ..., or 1:8). We set the batch size to 64 and the hyperparameter of LSR to 0.8 which we get by repetitive parameter-adjustable experiments.

\subsection{Experimental Results}










For image-transforms based data augmentation methods, both the geometric and radiometric transformation operations are utilized for comparison. Table \ref{table2} gives the comparison results. As can be seen in Table \ref{table2}, as the ratio of augmented/original samples increases, the classification overall accuracies obtained by these methods show a gradual upward trend. But radiometric transformation operations based methods are extremely weak in accuracy improvement, compared with the geometric based methods. This indicates that the samples produced by methods that using \emph{Image Enhancement} and \emph{Additive Noise} operations have no use for improving the remote sensing scene classification accuracies, which may result from the reason that these radiometric transformation operations do not enrich the sample variety in terms of spatial information, a factor playing a more important role in scene classification than color information. All the geometric based methods perform much better than the radiometric based methods, which demonstrates that the geometric-transforms samples are more effective than the radiometric-transforms samples on scene classification. As for the geometric based methods, the performances of two operations, \emph{i.e. Flip and Rotation}, are on equal terms. However, these operations are extremely limited in quantity of transforms samples, largely confined by avaliable transforming modes. 
For example, in the condition of retaining all the information in original images, the Rotation and Flip can only output threefold and fourfold fresh images respectively.
In general, the geometric transforms can bring the improvement while the radiometric transforms cannot. Therefore, in the following experiments, a combination of \emph{Flip} and \emph{Rotation} operations is applied for traditional image-transforms based data augmentation method, which is compared with the proposed SiftingGAN method. 

\begin{table}
\caption{Classification accuracies(\%) in the form of Means $\pm$ Standard Deviation with different data ratios of traditional image-transforms based data augmentation methods. The results show that the Geometric Transforms can bring the improvement while the Radiometric Transforms cannot.}
\label{table2}
\setlength{\tabcolsep}{1mm}{
\begin{tabular}{c|c|c|c|c}
\hline
\multirow{2}{*}{Methods} & \multirow{2}{*}{Operations} & \multicolumn{3}{c}{ Ratios (original:augmented)}\\
\cline{3-5}
& & 1:1 & 1:2 & 1:3 \\
\hline
\multirow{2}{*}{Geometric}	& \textbf{Flip}&	$90.64 \pm 0.47$ & $91.46 \pm 0.32$ & $91.81 \pm 0.63$ \\
	& \textbf{Rotation}	&	$90.33 \pm 0.55$ & $91.49 \pm 0.61$ & $92.09 \pm 0.56$ \\
\hline
\multirow{2}{*}{Radiometric} & Enhancement	&	$89.42 \pm 0.26$ & $89.74 \pm 0.29$ & $89.75 \pm 0.36$ \\
 & Noise	&	$89.24 \pm 0.33$ & $89.19 \pm 0.58$ & $89.54 \pm 0.28$ \\
\hline
\end{tabular}}
\label{tab2}
\end{table}

\begin{figure}[!t]
\centerline{\includegraphics[width=\columnwidth]{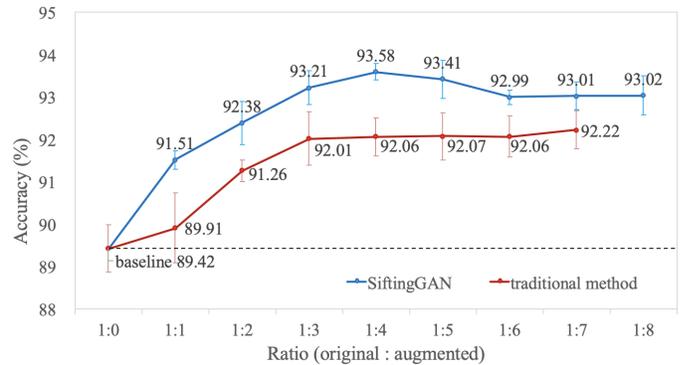}}
\caption{Comparison of classification accuracies between the traditional method (red line) and the proposed SiftingGAN (blue line) for data augmentation.}
\label{fig5}
\end{figure}

To further evaluate the effectiveness between the traditional method and our method for improving scene classification performance, Fig. \ref{fig5} displays and compares the classification accuracies of the traditional image-transforms based method and the proposed SiftingGAN for data augmentation. Firstly, it can be observed that the remote sensing scene classification baseline performance of fine-tuned VGGNet16 achieves an overall accuracy of 89.42\% which basically conforms to that in \cite{Xia2017AID}. When data augmentation is applied, both two methods are able to improve the baseline. The best improvements attained by SiftingGAN and the traditional method are +4.16\% and +2.80\% respectively, which demonstrates that data augmentation is capable of bringing improvement to the classification performance of DCNN. On the one hand, as the ratio of augmented/original samples increases, both two methods show rising trends. The accuracy achieved by our method improves faster and is significantly higher than that done by traditional method under the same ratio.  On the other hand, the performance of traditional method quickly saturates at the ratio of 1:3 while that of our method achieves its peak at 1:4. All these results indicate that the samples generated by our method contain more newly generated spatial information and are more helpful for training a DCNN model adequately. When excessive SiftingGAN-generated samples are utilized, the rising trend turns to slightly decrease, but the performance is still conspicuously higher than that of the traditional method. Note that the accuracy curve line terminates at 1:7 because of the extremely limited geometric transforming modes.

\section{Conclusion}
\label{sec:Conclusion}
In this paper, we extend the GAN framework and propose a SiftingGAN that can generate and sift labeled samples for data augmentation to improve remote sensing image classification baseline in other deep learning frameworks out of GANs. Experimental results on AID dataset verify that the proposed SiftingGAN method can not only effectively improve the performance of the scene classification baseline that is achieved without data augmentation, but also significantly excels the comparison methods based on traditional image transformation operations.
In the future, we will continue to investigate on whether the proposed SiftingGAN can be used on other data types such as hyperspectral images and synthetic aperture radar images. And we will apply the SiftingGAN to other GAN variants and prove its generalization.


\bibliographystyle{IEEEtran}
\bibliography{IEEEabrv,ReferenceAbrv}
\end{document}